\documentclass{article}

\usepackage{amssymb}

\usepackage[numbers]{natbib}
\usepackage{amsmath,amsthm}
\usepackage{graphicx}
\usepackage{algorithm}
\usepackage{algorithmic}
\usepackage{booktabs}
\usepackage{enumerate}
\usepackage{multirow}

\newtheorem{theorem}{Theorem}

\newtheorem{assumption}{Assumption}
\newtheorem{lemma}{Lemma}
\newtheorem{corollary}{Corollary}

\usepackage[letterpaper,top=2cm,bottom=2cm,left=3cm,right=3cm,marginparwidth=1.75cm]{geometry}

\title{Beyond Non-Expert Demonstrations: Outcome-Driven Action Constraint for Offline Reinforcement Learning}

\author{Ke Jiang\footnote{College of Computer Science and Technology, Nanjing University of Aeronautics and Astronautics, MIIT Key Laboratory of Pattern Analysis and Machine Intelligence. Email Address: \{ke\_jiang,x.tan\}@nuaa.edu.cn}, Wen Jiang$^*$, Yao Li\footnote{School of Computer and Information Technology, Shanxi University}, Xiaoyang Tan$^*$}

\begin{document}
\maketitle

\begin{abstract}
We address the challenge of offline reinforcement learning using realistic data, specifically non-expert data collected through sub-optimal behavior policies. Under such circumstance, the learned policy must be safe enough to manage \textit{distribution shift} while maintaining sufficient flexibility to deal with non-expert (bad) demonstrations from offline data.
To tackle this issue, we introduce a novel method called Outcome-Driven Action Flexibility (ODAF), which seeks to reduce reliance on the empirical action distribution of the behavior policy, hence reducing the negative impact of those bad demonstrations.
To be specific, a new conservative reward mechanism is developed to deal with {\it distribution shift} by evaluating actions according to whether their outcomes meet safety requirements - remaining within the state support area, rather than solely depending on the actions' likelihood based on offline data.
Besides theoretical justification, we provide empirical evidence on widely used MuJoCo and various maze benchmarks, demonstrating that our ODAF method, implemented using uncertainty quantification techniques, effectively tolerates unseen transitions for improved "trajectory stitching," while enhancing the agent's ability to learn from realistic non-expert data.

\paragraph{Keywords:}Offline Reinforcement Learning, Non-Expert Data, Outcome-Driven Constraint, Trajectory Stitching
\end{abstract}

\section{Introduction}

Offline reinforcement learning (RL)~\cite{li2022alleviating,li2023self} aims to learn a high-capacity policy from an offline dataset previously collected via a behavior policy~\cite{zhangzhe}, which has yielded significant improvements in various fields, including robotics tasks~\cite{mnih2015human,peng2017deeploco}, healthcare~\cite{li2024temporal,li2025novel}, game playing~\cite{silver2017mastering}, and large language models~\cite{achiam2023gpt,touvron2023llama}. However, prior studies~\cite{BCQ,cql} have indicated that offline RL algorithms suffered from the \textit{distributional shift}~\cite{BCQ,pessimism,pang2025qfae} problem, where the divergence between the new and behavior policies makes the agent encounter with some unseen actions or states~\cite{sdc,osr}, which are challenging for generalization during practical deployment.

Another significant problem in practice is that it is commonly expensive and challenging to obtain ideal expert data, and the realistic data used for training is usually generated through sub-optimal behavior policies, which means that the offline dataset contains lots of non-expert (bad) demonstrations. This would compound with the aforementioned \textit{distributional shift} problem and become more pronounced when learning from such realistic non-expert data, as blindly cloning these potentially highly sub-optimal behaviors can be harmful for policy improvement under this situation. For instance, many previous works, such as Behavior Regularized Actor-Critic (BRAC) \cite{brac}, Conservative Q-Learning (CQL) \cite{cql}, and TD3+BC \cite{td3bc}, focus on cloning expert behaviors and may be adversely affected by the sub-optimal behaviors present in the dataset \cite{pbrl}.
While more recent action-based support set approaches, such as Bootstrapping Error Accumulation Reduction (BEAR) \cite{BEAR}, Supported Policy Optimization (SPOT) \cite{SPOT} and Supported Value Regularization (SVR)~\cite{svr}, attempt to relax cloning conditions through supported regularization, they still face the challenge of being overly restrictive when learning from non-expert offline data. Specifically, they may suppress the likelihood of selecting actions those never taken by the behavior policy, i.e., OOD actions, including those unseen but generalizable actions, whose outcomes are in-distributional and safe.

\begin{figure}[t]
\centering
\includegraphics[width=1\linewidth]{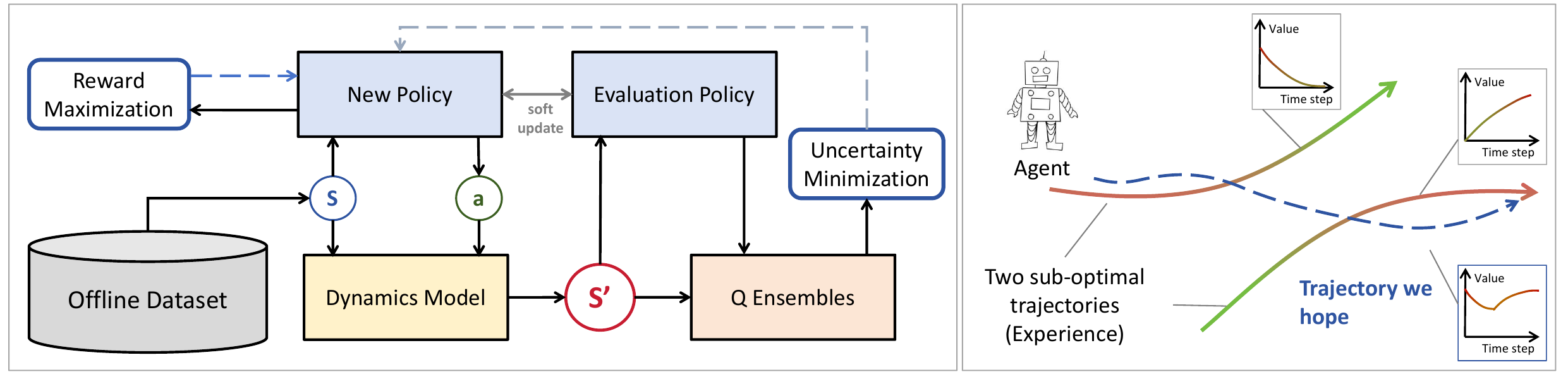}
\caption{The main framework (left) and basic idea (right) of our method. The left part is the training process of the ODAF. The right part is the process of trajectory stitching, where the agent stitches the high-value parts from different sub-optimal trajectories from offline data and generates a trajectory with higher value.}
\label{fig:ODAF_trajectorystitching}
\end{figure}

In this paper, we proposed a new method to address the above issues, whose main framework is shown in Figure \ref{fig:ODAF_trajectorystitching} (left). The key idea is to design a reward mechanism based on whether the subsequent states by following the current policy are: 1) beneficial to improve the performance, i.e., normal RL objective; 2) satisfying the safety requirements - falling within the state support area, instead of explicitly restricting the range of action space for each state. In other words, in our method, the OOD actions are \textit{allowed} as long as their consequences are safe and are beneficial to performance improvement. In this way, our approach is not only more conducive to shaping the desired behavior but also less susceptible to being misled by sub-optimal behavior policies. This is in contrast with the aforementioned action-support constraints-based offline reinforcement learning algorithms (e.g., SPOT, SVR, et al.), which overlooks the correlation between agent decision-making and potential outcomes, thus diminishing the agent’s flexibility in decision-making. 

In summary, our method focuses on the potential consequences that an action can yield, rather than the specific properties of the action itself, e.g., whether it looks like samples of the behavior policy. Actually, there are previous methods that can be seen through this lens. For example, State Deviation Correction (SDC)~\cite{sdc} and Out-of-sample Situation Recovery (OSR)~\cite{osr}, which were initially developed to help agents recover from Out-of-distribution (OOD) situations by trying to align the transition behavior of the learned policy with that of the behavior policy, can be thought of as matching the consequences of the actions with those in the dataset. However, even these methods are not robust against non-expert data as they do not take the quality of decisions' consequences into account. Actually, blindly cloning the transitions in the dataset may hinder the process of "trajectory stitching" in the case of a sub-optimal behavior policy. As illustrated in Figure \ref{fig:ODAF_trajectorystitching} (right), our proposed method, called \textbf{O}utcome-\textbf{D}riven \textbf{A}ction \textbf{F}lexibility (\textbf{ODAF}), naturally tolerates unseen transitions during the "trajectory stitching" process, thereby enhancing the agent's ability to learn from non-expert data, while such optimal trajectories cannot be synthesized by either traditional methods due to their reliance for the certain action distribution of behavior policy.

In what follows, after an introduction and a review of related works, Section \ref{sec:preliminary} provides a brief overview of the preliminary knowledge on basic formulation and action-supported methods in offline RL. Section \ref{sec:method} details the ODAF method, along with a theoretical explanation of its effect and the practical implementation. Experimental results are presented in Section \ref{sec:exper} to evaluate the effectiveness of both methods under various settings. Finally, the paper concludes with a summary.

\section{Related Works}


\subsection{Action-supported offline RL} Action-supported regularization plays a pivotal role in offline RL, striking a balance between conservatism and the ability of the learned policy to stitch trajectories. The Bootstrapping Error Accumulation Reduction (BEAR)~\cite{BEAR} method pre-trains an empirical behavior policy and regulates the divergence within a relaxation factor of the new policy. Supported Policy Optimization (SPOT)~\cite{SPOT} takes a different approach by explicitly estimating the behavior policy’s density using a high-capacity Conditional VAE (CVAE)~\cite{cvae} architecture. The most recent advancement in this field is Supported Value Regularization (SVR)~\cite{svr}, which simplifies action-supported regularization by only requiring an estimation of the behavior policy’s action visitation frequency, significantly reducing estimation errors and enhancing robustness. However, action-supported regularization would be too restrictive in avoiding all unseen actions, even those with safe consequences and are worth exploring.


\subsection{State recovery-based offline RL} As the relaxation of action-constraint, state recovery-based methods like State Deviation Correction (SDC)~\cite{sdc} align the transitioned distributions of the new policy and the behavior policy, forming a robust transition to avoid the OOD consequences. To further avoid the explicit estimation of consequences in high-dimensional state space, Out-of-sample Situation Recovery (OSR)~\cite{osr} introduces an inverse dynamics model (IDM)~\cite{markovrep} to consider the consequential knowledge in an implicit way when decision making. Besides, recent~\cite{mao2024offline} also considers the value of OOD consequences to enhance the performance in OOD state correction. In this paper, we also consider them as the Outcome-Driven methods that implicitly avoid the {\it state distributional shift} problem via aligning the transitioned distribution of the new policy with that of the behavior policy. But such methods would hinder their ability of {\it trajectory stitching} and generalization on non-expert data.


\subsection{Trajectory stitching in offline RL} 
Recently, Model-based Return-conditioned Supervised Learning (MBRCSL)~\cite{zhou2023free} is proposed to equip the agent with trajectory stitching ability. Although this method has achieved great improvement in certain scenarios, demonstrating the importance of {\it trajectory stitching}, it needs a large number of rollouts with the pre-trained model to correct the sub-optimal data distribution of the dataset, accumulating the model error. This motivates us to propose the ODAF method to achieve the {\it trajectory stitching} ability via only policy constraint.


\section{Preliminaries}\label{sec:preliminary}

A reinforcement learning problem is usually modeled as a Markov Decision Process (MDP), which can be represented by a tuple of the form $(S,A,P,R,\gamma, \rho_0)$, where $S$ is the state space, $A$ is the action space, $P$ is the transition probability matrix, $R$ and $\gamma$ are the reward function and the discount factor, $\rho_0$ is the initial state distribution. A policy is defined as $\pi:S\rightarrow A$ that makes decisions acting with the environment. 

In general, we define a Q-value function $Q^\pi(s,a) = (1 - \gamma)\mathbb E [\sum_{t=0}^\infty\gamma^t R(s_t,\pi(a_t|s_t))|s,a]$ to represent the expected cumulative rewards. Besides, we define the advantage as $A^\pi(s,a) = Q^\pi(s,a) - V^\pi(s)$, where $V^\pi(s) = \mathbb E_{a\sim \pi(\cdot|s)}[Q^\pi(s,a)]$. Then we define the $\gamma$-discounted future state distribution (stationary state distribution) for convenience as, $d^\pi(s) = (1-\gamma)\sum^{\infty}_{t=0}\gamma^t Pr(s_t=s; \pi,\rho_0)$, where $\rho_0$ is the initial state distribution and the $(1-\gamma)$ is the normalization factor.

In offline setting, Q-Learning~\cite{ql} learns a Q-value function $Q(s,a)$ and a policy $\pi$ from a dataset $\mathcal D$ collected by a behavior policy $\pi_\beta$, which consists of quadruples $(s,a,r,s')\sim d^{\pi_\beta}(s)\pi_\beta(a|s)P(r|s,a)P(s'|s,a)$. Then the objective is minimizing the Bellman error over the offline dataset~\cite{ql}, using exact or an approximate maximization scheme, such as CEM~\cite{cem}, onto the above method to recover the greedy policy, as follows:
	\begin{align}
	    	\min_Q &\mathbb E_{(s,a,r,s')\sim\mathcal D} \big[r + \gamma \mathbb E_{s'\sim P(s'|s,a)}[\mathbb E_{a'\sim\pi(a'|s')}Q'(s',a')] - Q(s,a) \big]^2
      \label{eq:qlearning}\\
                \max\limits_\pi &\mathbb E_{s\sim\mathcal D}\mathbb E_{a\sim\pi(\cdot|s)}[Q(s,a)]\label{eq:qlearning_pi}
	\end{align}
where $Q'$ is the target Q-value network.

\subsection{Action-supported offline RL}
 The well-known extrapolation error problem would occur~\cite{BCQ} when estimating the $\max_{a'}Q(s',a')$ in the Eq.(\ref{eq:qlearning})'s TD target. Methods, such as BEAR~\cite{BEAR}, SPOT~\cite{SPOT} and SVR~\cite{svr}, are proposed to address such issue while preserving the ability of trajectory stitching through the action-supported regularization. In general, these methods could be represented in the following form,
	\begin{align}
	    	&\min_Q \mathbb E_{(s,a,r,s')\sim\mathcal D} \big[r + \gamma \mathbb E_{s'\sim P(s'|s,a)}[\max_{\pi\in\Pi_{ac}}\mathbb E_{a'\sim\pi(\cdot|s')}Q(s',a')] - Q(s,a) \big]^2\label{eq:bear}\\
      &\quad\quad\quad\quad\quad\Pi_{ac} = \{\pi|\forall s, supp(\pi(a|s)) \subseteq supp(\pi_\beta(a|s))\}
      \label{eq:a_support_set}
	\end{align}
where the $\Pi_{ac}$ is the candidate policy set, in which all the policies $\pi$ would only generate actions supported by the behavior policy $\pi_\beta$.


\section{The Method} \label{sec:method}
In this section, we introduce the proposed method in detail. First, the objective of the proposed Outcome-Driven Action Flexibility (ODAF) is given in Sec.\ref{sec:odpb}. Then in Sec.\ref{sec:imple}, we give the way of implicit implementation, where we utilize the uncertainty lower bound of Q ensembles to approximate the ODAF constraint, which is utilized for empirical analysis in Sec.\ref{sec:exper}. Finally, the properties of the proposed method are discussed in a theoretical way in Sec.\ref{sec:theoreticalanalysis}.



\subsection{Outcome-Driven Policy Bootstrapping}\label{sec:odpb}
First, like the previous action-support methods in Eq.(\ref{eq:a_support_set}), we define an outcome-driven candidate set for policy search to regularize the consequences of the candidate policies to fall within the support set of offline data. So we can select the optimal policy that has highest expected returns from this set by a policy-constrained bootstrapping. This is more beneficial for balancing the performance and safety. The outcome-driven candidate set could be formulated as,
    \begin{align}
        \Pi = & \{\pi| \forall s\in\mathcal{D}, supp(P(s'|s, \pi)) \subseteq supp(d^{\pi_\beta}(s'))\}\label{eq:d_support_set}
    \end{align}

where $supp(p)$ denotes the support set of a distribution $p$, $d^{\pi_\beta}$ denotes the stationary state distribution of the behavior policy $\pi_\beta$, and $P(s'|s,\pi) = \mathbb E_{a\sim\pi(\cdot|s)}P(s'|s,a)$ is the transitioned distribution of the new policy $\pi$. In words, this $\Pi$ defines a policy set for a given environment based on some behavior policy $\pi_\beta$, in which each policy is safe in the sense that by following it, the transition state will always fall within the support of $d^{\pi_\beta}$. Comparing to previous methods, e.g., those defined in E.q.(\ref{eq:a_support_set}), we see that our candidate policy set $\Pi$ are based on the outcome of the policy rather than the behaviors the policy performed.

However, finding an optimal solution from the policy set defined in Eq. (\ref{eq:d_support_set}) is a computationally challenging problem. In what follows in this section, we will construct a formal value iterative framework to address this issue. Specifically, we first define the Outcome-Driven bootstrapping Bellman operator based on the constructed $\Pi$ as follows:
\begin{align}
        \hat{T}^{\Pi}Q(s,a) := r(s,a) + \gamma\mathbb E_{s'\sim \hat{P}(s'|s, a)} \max_{\pi\in\Pi} \mathbb E_{a'\sim\pi(a'|s')}Q(s',a')\label{eq:support_bellman}
    \end{align}
where $\hat{P}$ is the empirical dynamics model of the dataset. In particular, if using the true dynamics model $P$ to replace $\hat{P}$, the $\hat{T}^\Pi$ would be noted as $T^\Pi$. 
Comparing to traditional optimal Bellman operator as in Eq.(\ref{eq:qlearning}), such operator's TD target is estimated only through the optimal actions whose consequences are fully supported by the offline dataset. This makes the learned Q function not likely to overestimate on those unsafe actions, hence enhancing the safety on learning from offline data. On the other hand, deviating from those action-supported methods as in Eq.(\ref{eq:a_support_set}), our method does not rely on specific action distributions from the offline dataset, so it has a much better potential to generalize on those non-expert datasets.

The properties of Outcome-Driven bootstrapping Bellman operator, such as contraction and convergence, are discussed in Section \ref{sec:theoreticalanalysis} after the introduction for a practical implementation of the above idea.

\subsection{An Uncertainty-based regularization Algorithm for Implementation}\label{sec:imple}

To construct the outcome-driven candidate set $\Pi$ as defined in Eq.(\ref{eq:d_support_set}), we can utilize an $\epsilon-$approximation for the policy support set $\Pi\approx\Pi_\epsilon$, where $\Pi_\epsilon = \{\pi| \forall s\in\mathcal{D}, \forall s'\notin supp(d^{\pi_\beta}(s')), P(s'|s,\pi)<\epsilon\}$. This is a common way to deal with the support-based regularization that widely utilized~\cite{SPOT,svr}, which is also considered as a trick to enhance generalization in practical employment.
Then with a Lagrange approximation performed on the regularization $\pi\in\Pi_\epsilon$, we have the following objective function as a regularization for the new policy,
    \begin{align}
        \min_\pi\beta_{ssb}\cdot\sum_{s'\notin supp(d^{\pi_\beta}(s'))}(P(s'|s, \pi) - \epsilon)\label{eq:minsupport}
    \end{align}
where $\beta_{ssb}$ is the balancing coefficient. Then to implement Eq.(\ref{eq:minsupport}), recall that if we execute an action $a$ at any state $s$ in the offline dataset $\mathcal D$, the distribution of the potential consequence $s'$ is given by the dynamics model $P(s'|s, a)$. Our key idea is then to approximate Eq.(\ref{eq:minsupport}) from above based on the estimation of the state uncertainty of the state $s'$ resulted from a policy. We call it Outcome-Driven Action Flexibility (ODAF).


To this end, we may define a new learning objective by minimizing the state uncertainty of the new policy $\pi$ over the perturbed $s$, as follows,
\begin{align}
    \min_{\pi} \mathbb E_{s\sim\mathcal{D}}\left[\sum_{s'}P(s'|s, \pi)U^{\pi}(s')\right]
    \label{cumorleq}
\end{align}
where $U^{\pi}(s) = \mathbb E_{\pi(a|s)}U(s,a)$ and $U(s,a)\in[0,+\infty)$ is an uncertainty quantifier as is introduced in~\cite{ispessimism}, which has proven to have the property that if the input data $(s,a)$ is OOD, the $U(s,a)$ would be large and otherwise it would be small~\cite{edac}. Here we utilize it as the indicator to judge the averaged reliability of the learned policy $\pi$ over the potential consequences, aiming to margin out those behaviors which would lead to unsafe consequences.


Next we explicitly build the connection between the uncertainty-based regularization in Eq.(\ref{cumorleq}) and the support region of the dataset. In particular, Theorem \ref{theorem_cumorl} shows that 
under certain mild condition given in Assumption \ref{assumption_bounduncertainty}, we can use uncertainty to bound the support region of the dataset. 

\begin{assumption}\label{assumption_bounduncertainty}
    \textbf{(Bounded uncertainty)}. For all unseen state-action pairs, their uncertain would be positive, i.e., $\forall (s,a)\notin supp(\mathcal D)$, $U_{min}\leq U(s,a)$, where the constant $U_{min}>0$.
\end{assumption}
where $\mathcal D$ is the offline dataset and $supp(D)$ is the support of $\mathcal D$. Assumption \ref{assumption_bounduncertainty} assumes that for any OOD state-action pair, the uncertainty estimator is strictly positive, which conform to the empirical results in~\cite{edac}. 


\begin{theorem}\label{theorem_cumorl}
    Given an arbitrary state $s$, a conservative policy $\pi$ and a state estimator $U^{\pi}$ based on the policy $\pi$. Then the minimizing the ODAF term in Eq.(\ref{cumorleq}), i.e.,
    \begin{align}
    \min_{\pi} \sum_{s'}P(s'|s, \pi)U^{\pi}(s')\nonumber
    \end{align}
    is equivalent to minimizing the upper bound of the following objective as in Eq.(\ref{eq:minsupport}),
    \begin{align}
        \sum_{s'\notin supp(d^{\pi_\beta}(s'))}P(s'|s, \pi)
    \end{align}
    where $supp(d^{\pi_\beta}(s'))$ is the support of the dataset.
\end{theorem}


In Eq.(\ref{cumorleq}), the uncertainty quantification $U^\pi$ could be seen as a indicator to recognize those OOD states, as assumed in Assumption \ref{assumption_bounduncertainty}, then it penalizes the new policy for all actions whose outcomes are beyond the support of dataset, which is explicitly defined in Eq.(\ref{eq:minsupport}). Despite the intuitive connection between the two objectives, we also provide \textit{the detailed Proof of Theorem \ref{theorem_cumorl} in \ref{proofoftheorem}. }

By Theorem \ref{theorem_cumorl} , we see that it is less likely for the agent to select the actions that would transit to those states outside the support region of the dataset, hence avoiding being misled by the sub-optimal behavior data, as what may happen for a naive behavior cloning algorithm. \footnote{Empirical evidences that ODAF term is adequate for our needs are also given in Section \ref{sec:validation}.}

\noindent\textbf{Implementation.} In practice, we implement the proposed Outcome-Driven Action Flexibility (ODAF) onto a SAC-N~\cite{edac} framework. The ODAF in Eq.(\ref{cumorleq}) could be implemented as the loss,
\begin{align}
    L_{odaf} &= \mathbb E_{s\sim\mathcal D}\Big[\max_{\hat{s}\in\mathbb B^{\epsilon_{odaf}}_s}\big[\sum_{\hat{s}'}P(\hat{s}'|\hat{s}, \pi)U^{\pi'}(\hat{s}')\big]\Big]
    \label{cumorlloss}
\end{align}
where $B^{\epsilon_{odaf}}_s$ is a perturbation ball around state $s$ with magnitude $\epsilon_{odaf}$. The learned policy $\pi'$ is soft-updated via the new policy $\pi$ in this implementation. Here we implicitly assume that the $\mathbb B^{\epsilon_{odaf}}_s$ term is related to $\pi$ in that the state $s$ is sampled from the latter's state occupancy $d^{\pi}(\cdot)$. In words, the new objective Eq.(\ref{cumorleq}) aims to find a robust policy $\pi$ that minimizes the maximum (worst) possibility of driving the agent to encounter unfamiliar regions.

Here we select the standard deviation based uncertainty estimator in~\cite{pbrl,edac}: 
\begin{align}
   U^{\pi}(s)\approx\beta\cdot Std(Q^k(s,a)) = \beta\cdot \sqrt{\frac{1}{K}\sum_{k=1}^K\big(Q^k(s,a) - \bar{Q}(s,a)\big)}\label{eq:uncertainty_calcu}
\end{align}
where $\{Q^k\}_{k=1}^K$ is the Q-ensemble, $\bar{Q}$ is the averaged Q-value and $\beta$ is a constant. The Eq.(\ref{cumorlloss}) often utilize a Monte-Calro approximation in implementation.
Then we attach the $L_{odaf}$ in Eq.(\ref{cumorlloss}) onto the actor loss introduced in~\cite{pbrl} as,
	\begin{align}
                &L_{\pi}= -\mathbb E_{s\sim \mathcal D, a\sim \pi(\cdot|s)}\big[\min\limits_{j=1..N} Q'_j(s,a)-\beta \log \pi(a|s)\big] +\beta_{odaf}\cdot L_{odaf}
	\label{cumorlaeq}
	\end{align}
where $\beta_{odaf}$ is the wight of the ODAF term and $Q'$s are the target Q networks. The critic loss function $L_Q$ is as introduced in~\cite{pbrl},
	\begin{align}
                &L_{Q}= \mathbb E_{s,a,r,s'\sim \mathcal D} \big[\big(Q(s,a) - (r+ \gamma \mathbb E_{a'\sim\pi(\cdot|s')}[\min\limits_{j=1..N} Q'_j(s',a')-\beta \log \pi(a'|s')]\big)\big]
	\label{cumorlqeq}
	\end{align}
where $\{Q'_j\}_{j=1}^K$ are the $K$ Q ensembles.

To sum up, the whole ODAF could be implementation as Algorithm \ref{alg1}.

\begin{algorithm}[h]
    \caption{The pseudocode of the proposed Uncertainty-based Outcome-Driven Action Flexibility (ODAF) algorithm}
    \textbf{Input}: the offline dataset $\mathcal{D}$, maximal update iterations $T$, the pretrained dynamics model $P$. \\
    \textbf{Parameter}: policy network $\pi$, evaluation policy $\pi'$, Q-networks $Q_i$. \\
    \textbf{Policy Training}
    \begin{algorithmic} 
    \label{alg1}
        \STATE Initialize the policy network, Q-networks.
        \WHILE{$t < T$}
            \STATE Sample mini-batch of N samples $(s,a,r,s')$ from $\mathcal{D}$.
		  \STATE Get the perturbed $\hat{s}$ with adversarial method in~\cite{rorl}.
		  \STATE Feed $\hat{s}$ to the policy network $\pi$ and get $\hat{a}$.
            \STATE Feed $\hat{s}, \hat{a}$ to the dynamics model $P$ and get the potential consequence $\hat{s}'$.
		  \STATE Compute the agent's state uncertainty of $\hat{s}'$ according to the Q-networks $Q$ and $\pi'$.
		  \STATE Update the policy network $\pi$ according to Eq.(\ref{cumorlaeq}).
		  \STATE Update the Q-networks according to (\ref{cumorlqeq}).
            \STATE Soft update the parameters of the evaluation policy $\pi'$.
        \ENDWHILE
    \end{algorithmic}
        \textbf{Output: } The learned policy network $\pi$.
\end{algorithm}

\subsection{Theoretical Analysis}\label{sec:theoreticalanalysis}




In this section, we provide a theoretical analysis for the performance lower bound of the proposed method ODAF. First, Lemma \ref{lemma:contraction} shows that the Outcome-Driven bootstrapping Bellman operator is a contract operator in the Banach space. This is a critical to justify the convergence properties of the constructed Bellman operator.

\begin{lemma}\label{lemma:contraction}
    \textbf{(Contraction.)} The Bellman operator defined in Eq.(\ref{eq:support_bellman}) is a contraction operator.
\end{lemma}

\textit{The proof of Lemma \ref{lemma:contraction} is shown in Appendix \ref{appendix:theorem_main}.}

Then Theorem \ref{theorem:main} demonstrates the convergence property of the outcome policy of the constructed Outcome-Driven bootstrapping Bellman operator, under the assumption of the Outcome-Driven policy candidate set is well-learned according a constant $\epsilon$, i.e., $\forall \pi\in \Pi, s\in\mathcal{D}$, $\sup_{s'}\frac{\hat{P}(s'|s,\pi)}{d^{\pi_\beta}(s')}\leq \epsilon<1$. Such assumption is common in RL theory and aligned with assumptions in~\cite{xie2021bellman,cheng2022adversarially}.

\begin{theorem}\label{theorem:main}
    If we have constructed the outcome-driven policy candidate set $\Pi$ well, such that the transitioned distribution of all the candidate policies are covered by the dataset well, i.e., $\forall \pi\in \Pi, s\in\mathcal{D}$, $\sup_{s'}\frac{\hat{P}(s'|s,\pi)}{d^{\pi_\beta}(s')}\leq \epsilon<1$. Then we can bound the performance lower bound of our method,
    \begin{align}
        \|\hat{Q}^k - Q^*\|(s,a) & \leq \frac{\gamma R_{max}}{1-\gamma}\sqrt{\frac{2}{N\cdot d^{\pi_\beta}(s,a)}\log(\frac{|S||A|\cdot 2^{|S|}}{\delta})} + \gamma^k\cdot\epsilon\cdot\|\triangle_0\|\label{eq:main_theorem}
    \end{align}
    where $|S|,|A|$ are the dimensions of the state and action spaces. $\|\triangle_0\|=\max_{\pi\in\Pi}(\hat{Q}^0-\hat{Q}^*)$, where $\hat{Q}_0$ is an arbitrary initial value function and $\hat{Q}^*$ is the fixed point of $\hat{T}^\Pi$, and $Q^*$ is the fixed point of $T^\Pi$. $R_{max}$ is the upper bound of rewards and $N$ is the size of dataset.
\end{theorem}

\textit{The proof of Theorem \ref{theorem:main} is shown in Appendix \ref{appendix:theorem_main}. }

The results given by Theorem \ref{theorem:main} show that the distance between the learned Q function and the fixed point is bounded. However, Theorem \ref{theorem:main} does not guarantee the performance lower bound of the algorithm, which is commonly regarded as the distance between the outcome Q function and the true optimal Q function. Then Corollary \ref{corollary:theorem1} generalizes Theorem \ref{theorem:main} to the performance lower of the proposed ODFA method under certain condition.

\begin{corollary}\label{corollary:theorem1}
    If we assume the dataset has sufficient coverage over the optimal policy's stationary state distribution, i.e., $\sup_s\frac{d^{\pi^*}(s)}{d^{\pi_\beta}(s)}\leq C$, then Theorem \ref{theorem:main} can bound the learned agent's performance lower bound.
\end{corollary}

\begin{proof}
Firstly, from the optimal coverage assumption that $\sup_s\frac{d^{\pi^*}(s)}{d^{\pi_\beta}(s)}\leq C$, we can infer that all the consequences generated by the optimal policy $\pi^*$ would have a non-zero level of data coverage. Then the optimal policy is in the constructed outcome-driven policy candidate set, as defined in Eq.(\ref{eq:d_support_set}). Therefore, in Eq.(\ref{eq:support_bellman}), the maximum operation could always have the opportunity to select the actions generated by the optimal policy, and finally converge to the optimal value function. In other words, the fixed point of the constructed Bellman operator would be the true optimal value. Finally, the $Q^*$ in Theorem \ref{theorem:main} could be replaced by the true optimal Q function safely, and lower bound the performance of learnt policy.
\end{proof}

Theorem \ref{theorem:main} and Corollary \ref{corollary:theorem1} indicates that the performance of the value function learned by the Outcome-Driven bootstrapping Bellman operator constructed in Eq.(\ref{eq:d_support_set}) is influenced by the following aspects: 1) The size of dataset $N$. For the purpose of converging to the fixed point, the data number should be large, which is critical for the offline learning; 2) The step $k$. When the number of training iterations is large, or ideally to the infinity, the second term of Eq.(\ref{eq:main_theorem})'s right side would be tiny; 3) Dimensions of state and action spaces, i.e., $|S|,|A|$. These dimensions are commonly finite in practice, which means that the performance of the learned policy by our algorithm is successfully lower bounded. 



\section{Experiments}\label{sec:exper}

In experiments we mainly aim to answer the following three key questions:

\begin{enumerate}[1)]
\item Does ODAF achieve the state-of-the-art performance on standard MuJoCo benchmarks with non-expert data, compared to the latest closely related methods?\label{question:2}
\item Does ODAF has better stability (generalization ability) when learning on non-expert data?\label{question:3}
\item Does ODAF enable the agent to stitch the sub-optimal trajectories to achieve higher performance?\label{question:4}

\end{enumerate} 

Our experimental section is organized as follows: First, we verify that it is hard for the traditional methods to learn from non-expert datasets on standard MuJoCo benchmarks, but the proposed method ODAF has a superior performance among these methods, answering Question \ref{question:2}; Then, to answer Question \ref{question:3}, we perform ODAF in the MuJoCo with limited valuable data setting~\cite{sdc,osr,svr} to explore how the performance of ODAF changes when learning with different levels of non-expert data;
Also, we perform a test on PointMaze benchmark - a benchmark designed especially for testing the agent's {\it trajectory stitching}~\cite{zhou2023free} to directly confirm whether our method can achieve our claim - stitching for better trajectories and getting higher performance, answering Question \ref{question:4}; 
Besides, we also conduct the experiments over the more complicated tasks of AntMaze to evaluate the ability of multi-step dynamic programming; Evaluation on adversarial benchmarks are also performed to verify the robustness of our method;
Finally, we conduct the validation study and ablation study to verify what role the ODAF term plays. More experimental details (e.g. hyperparameters and network structures) could be found in \ref{externel_experiments}. Our code for an experimental demo is available at https://github.com/Jack10843/ODAF-master.

\begin{table*}[h]
\centering
\caption{Results of \textbf{ODAF(ours)}, CQL, PBRL, SPOT, SVR, EDAC, RORL, SDC and OSR-10 on offline MuJoCo tasks averaged over 4 seeds. We bold the highest scores in each task.}
\resizebox{0.95\columnwidth}{!}{
\begin{tabular}{ll|llllllll|ll}
\toprule
                             &     & CQL  & PBRL  & SPOT  & SVR  & EDAC  & RORL  & SDC   & OSR-10   & ODAF(Ours)\\ \midrule
\multirow{5}{*}{\rotatebox[
origin=c]{90}{halfcheetah}}  & r   & 17.5 & 11.0  & 26.5  & 27.2  & 28.4  & 28.5  & \textbf{36.2}  & 26.7   & 30.2±1.7   \\
                             & m   & 47.0 & 57.9  & 58.4  & 60.5  & 65.9  & 66.8  & 47.1  & 67.1    & \textbf{68.7±0.3}   \\
                             & m-e & 75.6 & 92.3  & 86.9  & 94.2  & 106.3 & 107.8 & 101.3 & 108.7  & \textbf{111.1±2.4}  \\
                             & m-r & 45.5 & 45.1  & 52.2  & 52.5  & 61.3  & 61.9  & 47.3  & 64.7    & \textbf{65.1±0.3}   \\
                             & e   & 96.3 & 92.4  & 97.6  & 96.1  & 106.8 & 105.2 & 106.6 & 106.3  & \textbf{107.9±1.1}\\ \hline
\multirow{5}{*}{\rotatebox[
origin=c]{90}{hopper}}       & r   & 7.9  & 26.8  & 28.7  & 31.0  & 25.3  & 31.4  & 10.6  & 30.4    & \textbf{32.1±1.5}   \\
                             & m   & 53.0 & 75.3  & 86.0  & 103.5 & 101.6 & 104.8 & 91.3  & 105.5  & \textbf{106.3±1.2}  \\
                             & m-e & 105.6& 110.8 & 99.3  & 111.2 & 110.7 & 112.7 & 112.9 & 113.2  & \textbf{114.3±0.8}  \\
                             & m-r & 88.7 & 100.6 & 100.2 & 103.7 & 101.0 & 102.8 & 48.2  & 103.1  & \textbf{104.8±0.8}  \\
                             & e   & 96.5 & 110.5 & 112.3 & 111.1 & 110.1 & 112.8 & 112.6 & 113.6  & \textbf{114.7±0.7}\\ \hline
\multirow{5}{*}{\rotatebox[
origin=c]{90}{walker2d}}     & r   & 5.1  & 8.1   & 5.0  & 2.2   & 16.6  & 21.4  & 14.3  & 19.7    & \textbf{24.4±2.3}   \\
                             & m   & 73.3 & 89.6  & 86.4  & 92.4  & 92.5  & 102.4 & 81.1  & 102.0  & \textbf{104.1±2.8}  \\
                             & m-e & 107.9& 110.8 & 112.0 & 109.3 & 114.7 & 121.2 & 105.3 & 123.4  & \textbf{123.8±0.7}  \\
                             & m-r & 81.8 & 77.7  & 91.6  & 95.6  & 87.1  & 90.4  & 30.3  & 93.8    & \textbf{95.1±1.9}   \\ 
                             & e   & 108.5& 108.3 & 109.7 & 110.0 & 115.1 & 115.4 & 108.3 & 115.3  & \textbf{115.9±1.3}\\ \hline
average                      &     & 67.4 & 74.4  & 76.9  & 80.0  & 82.9  & 85.7  & 70.2  & 86.2    & \textbf{87.9}       \\ \bottomrule
\end{tabular}}
\label{comparison}
\end{table*}

\subsection{Learning from non-expert datasets }

In this section, we compare our method with several significant methods, including CQL~\cite{cql}, PBRL~\cite{pbrl}, SPOT~\cite{SPOT}, SVR~\cite{svr}, EDAC~\cite{edac}, RORL~\cite{rorl}, SDC~\cite{sdc} and OSR-10~\cite{osr}, based on the D4RL~\cite{d4rl} dataset in the standard MuJoCo benchmarks. \textbf{MuJoCo (D4RL).} There are three types of high-dimensional control environments representing different robots in D4RL: Hopper, Halfcheetah and Walker2d, and five kinds of datasets: 'random', 'medium', 'medium-replay', 'medium-expert' and 'expert'. The 'random' is generated by a random policy and the 'medium' is collected by an early-stopped SAC~\cite{sac} policy. The 'medium-replay' collects the data in the replay buffer of the 'medium' policy. The 'expert' is produced by a completely trained SAC. The 'medium-expert' is a mixture of 'medium' and 'expert'.

The results is shown in Table \ref{comparison}, where part of the results for the comparative methods are obtained by~\cite{rorl,osr}. We have observed that the performance of all methods experiences a significant decrease when applied to non-expert datasets such as 'random', 'medium', 'medium-replay', and 'medium-expert'. This highlights the inherent difficulty in learning from non-expert data in practical settings. However, our proposed method, ODAF, consistently outperforms other approaches across most benchmarks, particularly surpassing methods that rely on behavior cloning such as CQL, PBRL, and EDAC. Furthermore, ODAF achieves state-of-the-art performance in terms of the average score. Additionally, we would like to emphasize that ODAF demonstrates significant improvements over the state-of-the-art conservative methods (e.g., SVR and OSR) on the 'medium' and 'medium-replay' datasets. This notable margin can be attributed to ODAF's ability to avoid error compounding through its flexibility in trajectory stitching. This further underscores the advantages of ODAF in effectively handling non-expert data. In the next section, we will delve deeper into exploring the advantages of ODAF across different levels of non-expert datasets.

\subsection{Influence of different levels of non-expert data}

In this section, we further explore the feasibility of the proposed ODAF on different levels of non-expert offline dataset, where we mix the 'expert' and 'random' datasets with different ratios. This is a setting widely used, such as in~\cite{sdc,svr,osr}. Here, the proportions of 'random' data are 0.5, 0.6, 0.7, 0.8 and 0.9, for Halfcheetah, Hopper and Walker2d.

\begin{figure*}[h]
    \centering
    \includegraphics[width=1\linewidth]{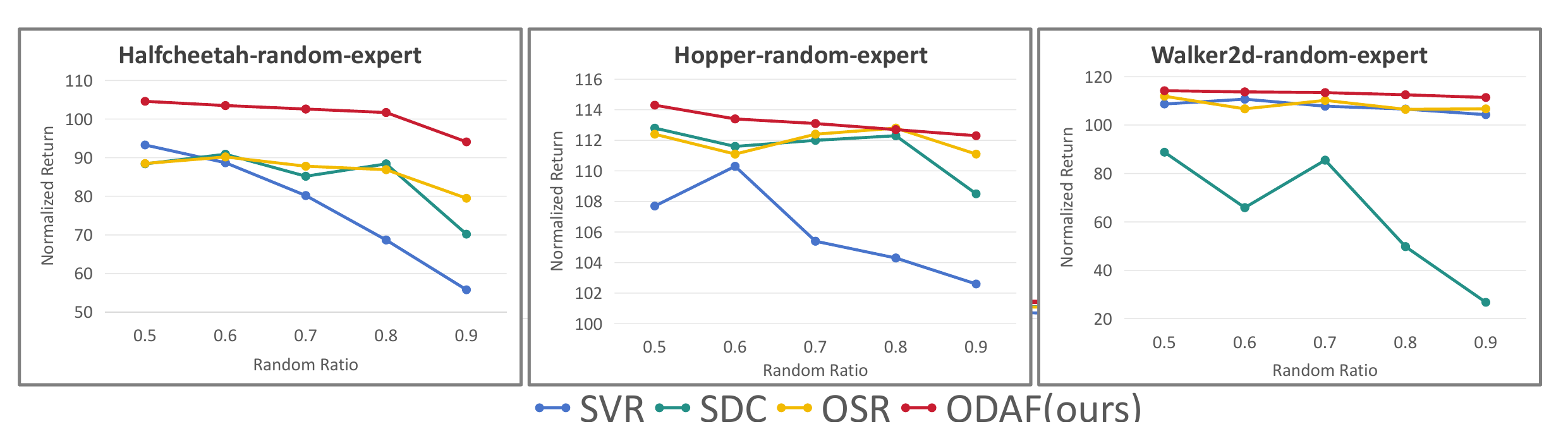}
    \caption{The results on the MuJoCo benchmarks with different levels of non-expert data.}
    \label{fig:limitedvaluable}
\end{figure*}


We compare the proposed ODAF with SVR~\cite{svr}, OSR~\cite{osr} and SDC~\cite{sdc}, as is shown in Figure \ref{fig:limitedvaluable}, ODAF outperforms the other three methods on three type of control environments over the normalized scores. We have observed that both of our proposed methods, particularly ODAF, exhibit a significantly lower decrease rate over the 'Halfcheetah' benchmark compared to the other two methods as the random ratio increases, which can be attributed to the agent's heightened sensitivity to the quality of data collection in this environment. Furthermore, when testing on the 'Hopper' and 'Walker2d' benchmarks, we note that ODAF demonstrates the least decrease in performance among all methods when the random ratio reaches 0.9, which highlights the advantage of the implicit implementation in addressing more complex tasks and learning from data of lower quality in practical scenarios. Therefore, we emphasize that our method, ODAF, is better equipped for learning with non-expert data, and they exhibit improved stability and performance across various benchmarks with lower data quality.

\subsection{PointMaze: Trajectory stitching testing}

\begin{figure}[h]
\centering
\includegraphics[width=0.6\linewidth]{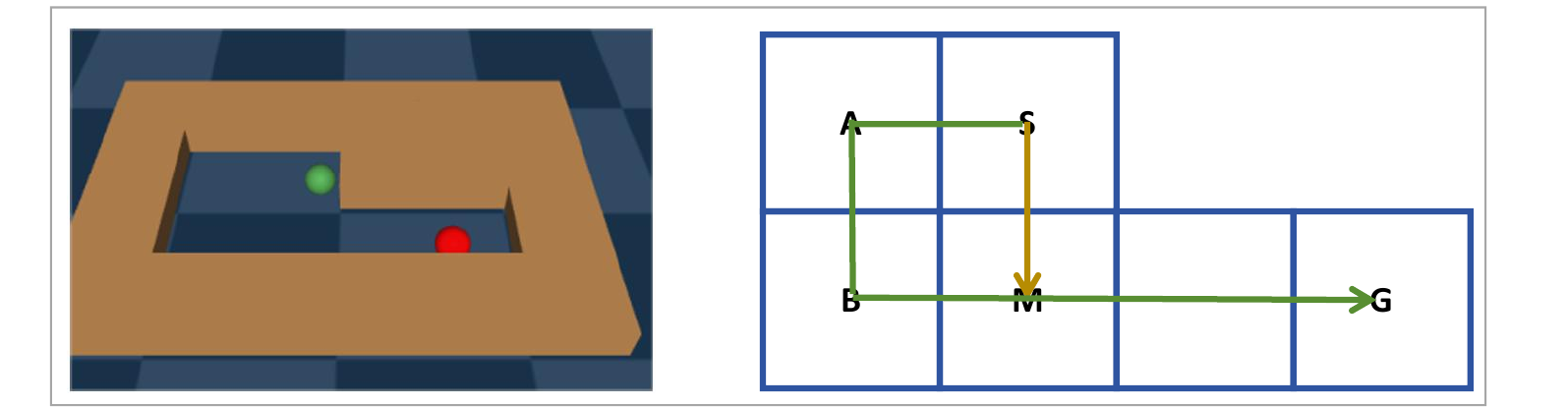}
\caption{The PointMaze map we used. The right shows the dataset description, where S is the initial point and G is the goal. The green line is a sub-optimal trajectory while the yellow line is a trajectory for stitching.}
\label{fig:exp_pointmaze}
\end{figure}

To investigate if the learned agents could do stitching, we introduce a specially designed PointMaze dataset~\cite{zhou2023free}, which consists of two kinds of sub-optimal trajectories with equal number, as is shown in (a) and (b) of Figure \ref{fig:exp_pointmaze}: 1) A detour trajectory S → A → B → G that reaches the goal in a sub-optimal manner; 2) A trajectory for stitching: S → M, whose return is very low, but is essential for getting the optimal policy. The optimal trajectory should be a stitching of the two trajectories in dataset (S → M → G).  \textbf{The resulting dataset has averaged return 40.7 and highest return 71.8}.

\begin{figure}[h]
\centering
\includegraphics[width=0.8\linewidth]{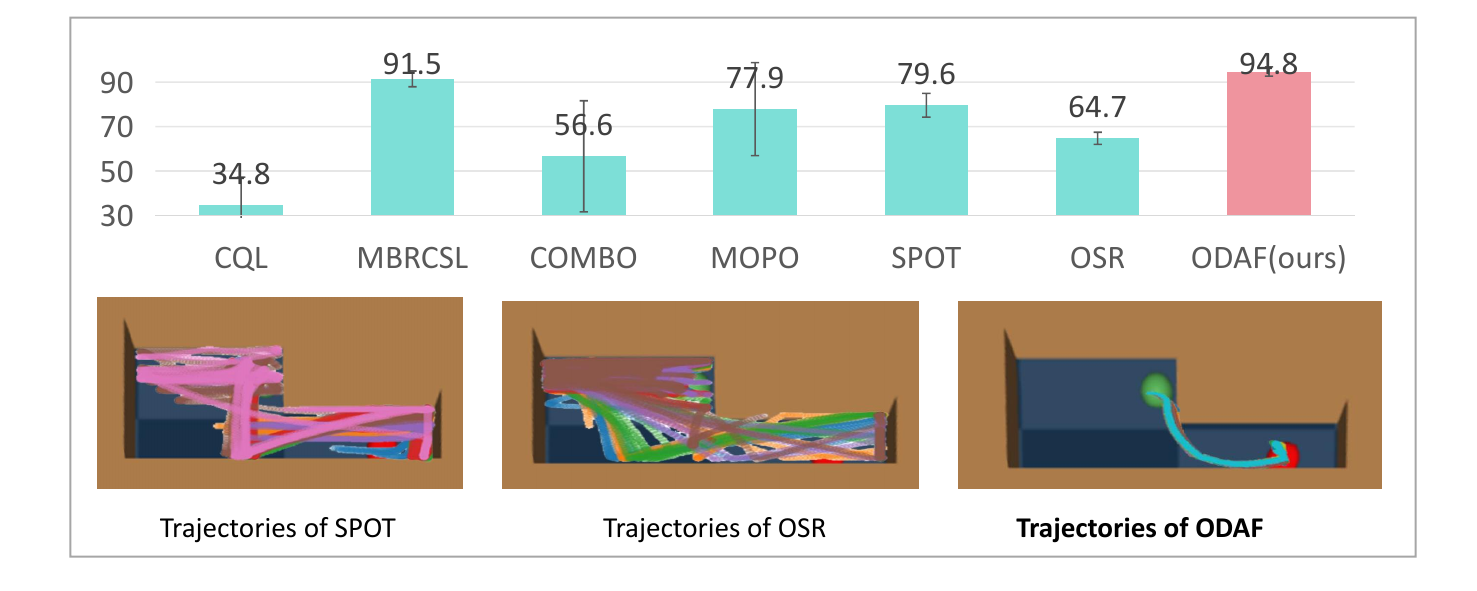}
\caption{The results of the methods. The proposed ODAF is marked red and the highest score is bolded. The bottom is the visualization of part of results in left.}
\label{fig:exp_pointmaze_res}
\end{figure}

To answer question \ref{question:4}, we compare the proposed ODAF with several offline RL baselines, including: 1) Traditional action-constraints: CQL~\cite{cql} and SPOT~\cite{SPOT}; 2) Outcome-Driven method: OSR~\cite{osr}; 3) Model-based methods: COMBO~\cite{yu2021combo} and MOPO~\cite{mopo}; 4) The method specially designed for trajectory stitching: MBRCSL~\cite{zhou2023free}. The results are shown in Figure \ref{fig:exp_pointmaze_res}, where the ODAF and MBRCSL both outperforms all the other baselines with a large margin, successfully stitching together sub-optimal trajectories. However, unlike MBRCSL, our method ODAF does not need large number of rollouts based on the approximated dynamics model, which means that ODAF is less likely to suffer from the error accumulation of the learned model, achieving higher performance and better efficiency.

In the bottom of Figure \ref{fig:exp_pointmaze_res}, we observe that the trajectories generated via SVR and OSR are scattered and coincide with the trajectories listed in the dataset, which demonstrates that these two methods would significantly over-fit to the transitions listed in the dataset instead of generalizing to those unseen but with higher value. However, the proposed ODAF successfully generate trajectories stitched with the two kinds of samples demonstrated in the dataset, achieving higher performance.

\subsection{Experiments on More Complicated Environment - AntMaze}
\begin{table*}[h]
\centering
\caption{Results of \textbf{ODAF(ours)}, CQL, IQL, SPOT, ATAC, SDC and OSR-10 on offline AntMaze tasks averaged over 4 seeds. We bold the highest scores in each task.}
\resizebox{0.95\columnwidth}{!}{
\begin{tabular}{ll|llllllll|ll}
\toprule
                             &     & CQL  & IQL  & SPOT  & ATAC  & SDC   & OSR-10   & ODAF(Ours)\\ \midrule
\multirow{5}{*}{\rotatebox[
origin=c]{90}{AntMaze}}      & umaze   & 82.6  & 87.5     & 93.5  & 70.6     & 89.0  & 89.9    & \textbf{94.6±0.9}   \\
                             & umaze-diverse & 10.2  & 62.2     & 40.7  & 54.3     & 57.3  & \textbf{74.0}   & 71.3±4.7  \\
                             & medium-play & 59.0  & 71.2     & 74.7  & 72.3     & 71.9  & 66.0   & \textbf{79.0±2.1}  \\
                             & medium-diverse & 46.6  & 70.0     & 79.1  & 68.7     & 78.7  & \textbf{80.0}    & \textbf{79.6±1.7}   \\ 
                             & large-play & 16.4  & 39.6     & 35.3  & 38.5     & 37.2  & 37.9   & \textbf{59.3±5.7}\\ 
                             & large-diverse & 3.2   & \textbf{47.5}     & 36.3  & 43.1     & 33.2  & 37.9   & \textbf{47.4±9.3} \\ \hline
average                      &     & 36.3  & 63.0     & 59.9  & 57.9     & 61.2  & 64.3    & \textbf{71.9}       \\ \bottomrule

\end{tabular}
}
\label{tab:antmaze}
\end{table*}

Compared to the MuJoCo environment, the AntMaze environment requires the agent to have the ability of multi-step dynamic planning, making it considered a more complex scenario. In this environment, we compare CQL~\cite{cql}, IQL~\cite{iql}, SPOT~\cite{SPOT}, ATAC~\cite{cheng2022adversarially}, SDC~\cite{sdc}, and OSR-10~\cite{osr}. In the AntMaze environment, based on the size and shape of the maze, it can be categorized into 'umaze,' 'medium,' and 'large'; and based on different tasks, it can be classified as 'diverse' and 'play'.
From the results in Table \ref{tab:antmaze}, we can observe that our method outperforms other methods in most environments, particularly in the 'large' and 'diverse' tasks, where our method significantly outperforms others. This indicates that our method exhibits strong generalization capabilities even when facing more complex and challenging tasks.

\subsection{Validation Study for ODAF Regularization}\label{sec:validation}

In this section, we perform a series of validation experiments to explore the impact of two key components of the proposed method: the pre-trained dynamics models \(\hat{P}(s'|s,a)\) and the uncertainty approximations \(U^\pi(s')\). Both components are integrated into the ODAF term in Eq. (\ref{cumorleq}), which evaluates the safety of the outcome resulting from a given action. To assess the effectiveness of the ODAF term, we conducted a straightforward experiment within the MuJoCo environment.

In the experiment, we first generated two sets of actions: one set with safe outcomes, obtained by selecting two similar states from the dataset and generating actions through the inverse dynamics model; the other set with unsafe outcomes, composed of a series of random actions. We then utilized either the true dynamics model (TDM) or our pre-trained dynamics model (PDM) to predict the next states of these actions and assess their safety as $
score(s,a) = \mathbb{E}_{\hat{P}(s'|s,a)} U^\pi(s').$


\begin{figure}[h]
\centering
\includegraphics[width=0.9\linewidth]{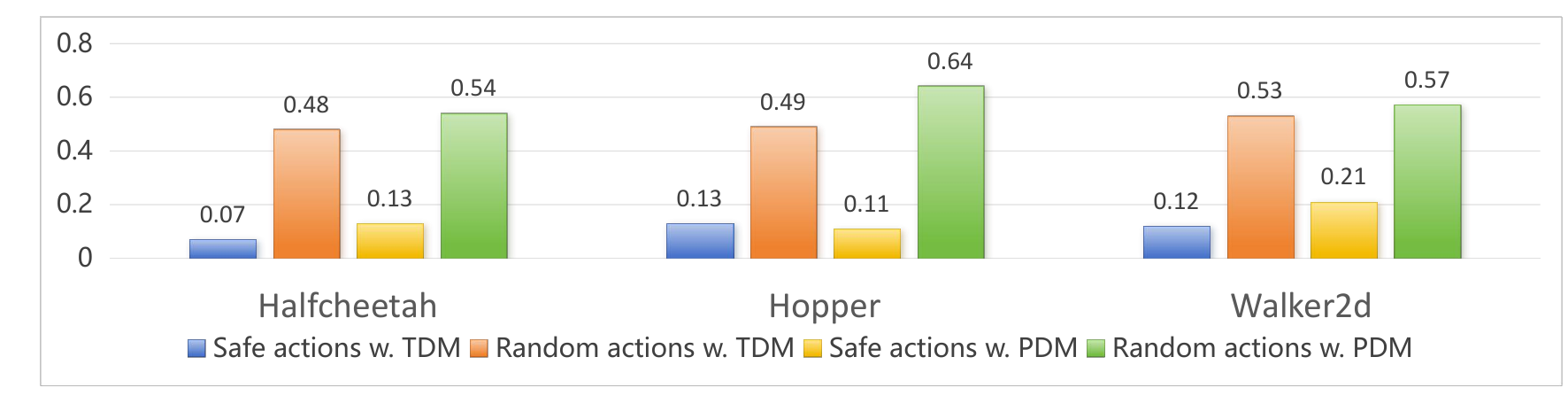}
\caption{Validation study for ODAF regularization.}
\label{tab:validation}
\end{figure}


Figure \ref{tab:validation} shows the results. Comparing the results of the blue and orange columns, we observe that our safety scoring method is sensitive to whether the consequences of actions are in-distribution or out-of-distribution (OOD), which supports the validity of this measurement. Looking at the results of the yellow and green columns, we find that the uncertainty quantifier also reveals a significant score gap between the two types of actions when using the pre-trained dynamics model. This gap is comparable to that observed in the first and second rows. This suggests that the performance of the pre-trained dynamics model is sufficient to distinguish whether the consequences of actions are safe, without even requiring a perfect reconstruction of the outcome state of those actions.

\subsection{Ablation study}

\begin{figure}[h]
\centering
\includegraphics[width=0.8\linewidth]{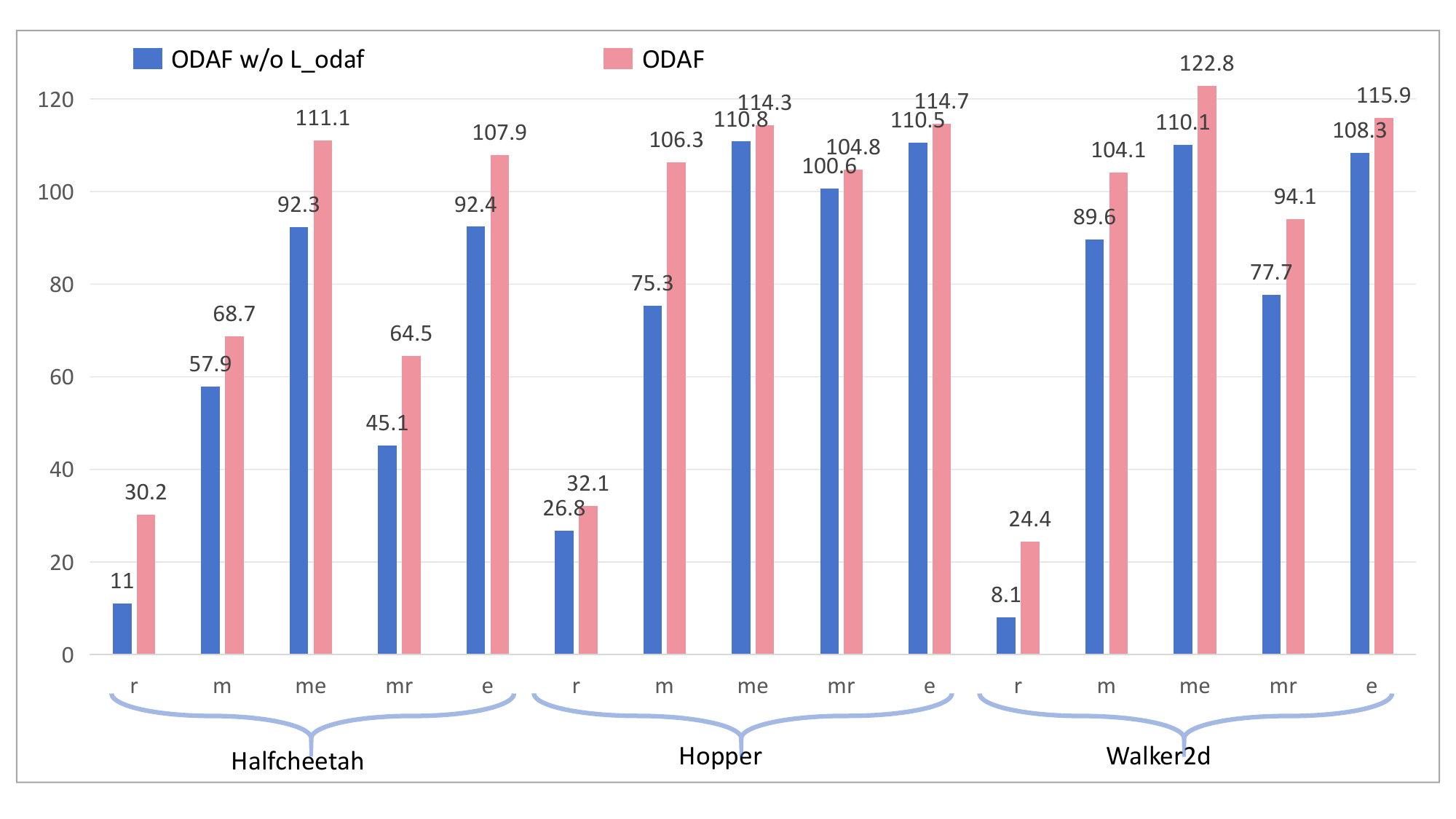}
\caption{The results of ODAF and ODAF w/o $L_{odaf}$ on the standard MuJoCo benchmarks.}
\label{ablation_figure}
\end{figure}

In this section, we perform an ablation study on the two implementations to evaluate how the ODAF term behaves. From the results in the left part of Figure \ref{ablation_figure}, we observe that ODAF significantly outperforms ODAF w/o $L_{odaf}$ (the regularization term in Eq.(\ref{cumorlaeq})) by nearly 20\% improvement on average, which demonstrates the important role ODAF term playS in learning a higher-capacity policy that is more likely to control the agent moving within the high-valuable regions.

\begin{figure}[h]
\centering
\includegraphics[width=1\linewidth]{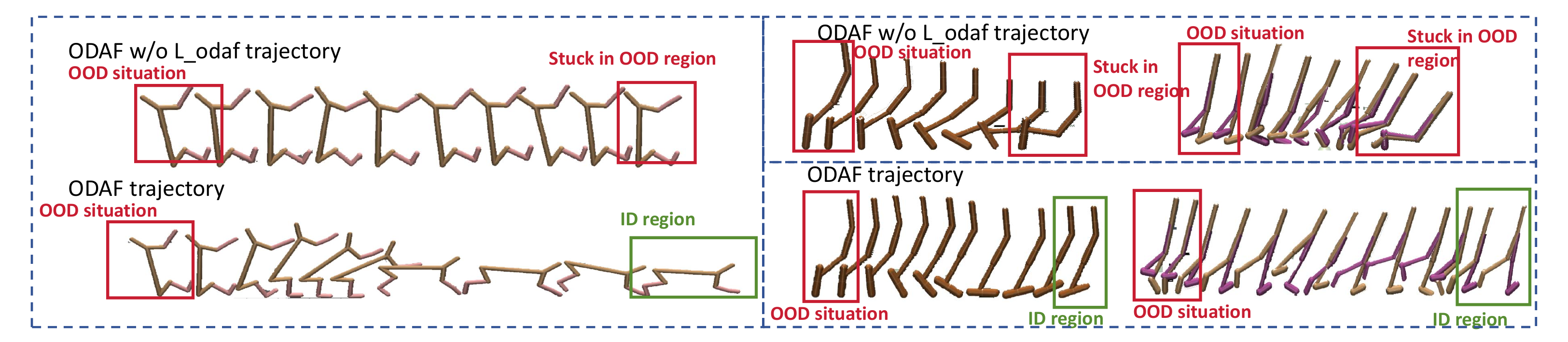}
\caption{The visualized results of ODAF and ODAF w/o $L_{odaf}$ at OOD samples.}
\label{ablation_vis}
\end{figure}

Furthermore, we also visualize some of the results of ODAF and ODAF w/o $L_{odaf}$ on the 'Halfcheetah-OOS-large', 'Hopper-OOS-large' and 'Walker2d-OOS-large' benchmarks, as shown in Figure \ref{ablation_vis}, from which we observe that compared with the results of ODAF w/o $L_{odaf}$, the ODAF agent generalizes better when falling into OOD situations and is more likely to generate transitions with those in-distributional consequences, enhancing the robustness, which could also be seen as a phenomenon that follows the (loosed) {\it state recovery} principle in another way.

\section{Conclusion}

In this paper, we propose a novel method called Outcome-Driven Action Flexibility (ODAF) to trade-off the conservatism and generalization when learning from non-expert data in offline RL. In particular, ODAF liberates the agent from the shackles of non-expert data in a Outcome-Driven manner - it implicitly avoids the agent suffering from {\it distributional shift} via controlling its consequences within in-distributional (safe) regions, while preserving its ability of trajectory stitching, which is critical for achieving superior performance from non-expert demonstrations. Theoretical and experimental results validate the effectiveness and feasibility of ODAF. We believe that the problem addressed in this work and the proposed method hold promise for practical applications of offline RL.

\section{Acknowledge}
This work is partially supported by National Natural Science Foundation of China (6247072715).

\appendix

\section{Proofs}

\subsection{Proof of Lemma \ref{lemma:contraction} and Theorem \ref{theorem:main}. }\label{appendix:theorem_main}

\textbf{Lemma \ref{lemma:contraction}. }\textbf{(Contraction.)} \textit{The Bellman operator defined in Eq.(\ref{eq:support_bellman}) is a contraction operator.}

\textit{Proof. }     Suppose there exist two variables $u,v$ in the value function space, then we have,
    \begin{align}
        \|T^\Pi u - T^\Pi v\|_\infty &= \max_s |T^\Pi u(s) - T^\Pi v(s)|\\
        & = \max_s |\max_{\pi\in\Pi}\mathbb E[r + \gamma u(s')|s,a] - \max_{\pi\in\Pi}\mathbb E[r + \gamma v(s')|s,a]|\\
        & \leq \max_s \max_{\pi\in\Pi}|\gamma\mathbb E[ u(s') - v(s')|s,a]|\\
        & \leq \max_s \max_{\pi\in\Pi}\gamma\mathbb E[ |u(s') - v(s')| \big|s,a]\\
        & \leq \max_s \max_{\pi\in\Pi}\gamma\mathbb E[ \|u(s') - v(s')\|_\infty \big|s,a]\\
        & = \gamma \|u(s') - v(s')\|_\infty
    \end{align}
    Completing the proof.

\textbf{Theorem \ref{theorem:main}. }\textit{If we have constructed the policy candidate set $\Pi$, such that the transitioned distribution of all the candidate policies are covered by the dataset well, i.e., $\forall \pi\in \Pi, s\in\mathcal{D}$, $\sup_{s'}\frac{\hat{P}(s'|s,\pi)}{d^{\pi_\beta}(s')}\leq \epsilon<1$. Then we can bound the performance lower bound of our method,}
    \begin{align}
        \|\hat{Q}^k - Q^*\|(s,a) & \leq \frac{\gamma R_{max}}{1-\gamma}\sqrt{\frac{2}{N\cdot d^{\pi_\beta}(s,a)}\log(\frac{|S||A|\cdot 2^{|S|}}{\delta})} + \gamma^k\cdot\epsilon\cdot\|\triangle_0\|
    \end{align}
\textit{where $|S|,|A|$ are the dimensions of the state and action spaces. $\|\triangle_0\|=\max_{\pi\in\Pi}(\hat{Q}^0-\hat{Q}^*)$, where $\hat{Q}_0$ is an arbitrary initial value function and $\hat{Q}^*$ is the fixed point of $\hat{T}^\Pi$, and $Q^*$ is the fixed point of $T^\Pi$. $R_{max}$ is the upper bound of rewards and $N$ is the size of dataset.}

\textit{Proof of Theorem \ref{theorem:main}.} 
First we decompose the $\|\hat{Q}^k - Q^*\|(s,a) = \|\hat{Q}^k - \hat{Q}^*\|(s,a) + \| \hat{Q}^* - Q^*\|(s,a)$ with the triangle inequality.

Then we aim to bound $\|\hat{Q}^* - Q^*\|$. First, by the triangle inequality, we have,
\begin{align}
    \|\hat{Q}^* - Q^*\|(s,a)\leq \|\hat{T}^\Pi \hat{Q}^* - \hat{T}^\Pi Q^*\|(s,a) + \|\hat{T}^\Pi Q^* - Q^*\|(s,a)
\end{align}
Because $\hat{T}^\Pi$ is a $\gamma$-contraction operator (see Lemma \ref{lemma:contraction}), we have,
\begin{align}
    \|\hat{Q}^* - Q^*\|(s,a)\leq & \frac{\|T^\Pi Q^* - T^\Pi Q^*\|(s,a)}{1-\gamma}\\
    = & \frac{\gamma}{1-\gamma}\|P(s'|s,a) - \hat{P}(s'|s,a)\|_1 \max_{\pi\in\Pi}Q^*(s',\pi)\\
    \leq & \frac{\gamma\cdot R_{max}}{1-\gamma}\|P(s'|s,a) - \hat{P}(s'|s,a)\|_1\\
    \leq & \frac{\gamma\cdot R_{max}}{1-\gamma}\sqrt{\frac{2}{N\cdot d^{\pi_\beta}(s,a)}\log(\frac{|S||A|\cdot 2^{|S|}}{\delta})}
\end{align}
The last inequality holds because of the \textbf{Proposition 9} in \cite{ghavamzadeh2016safe}.

Then we bound the $\|\hat{Q}^k - \hat{Q}^*\|(s,a)$. 
\begin{align}
    \|\hat{Q}^k - \hat{Q}^*\|(s,a) \leq^{(a)} & \gamma^k\|\mathbb E_{\hat{P}(s_k|\{\pi_1...\pi_{k-1}\}, s, a)}\max_{\pi\in\Pi}(\hat{Q}^0 - \hat{Q}^*)(s_k,\pi)\|\\
    = & \gamma^k\|\sum_{s_k}\frac{\hat{P}(s_k|\{\pi_1...\pi_{k-1}\}, s, a)}{d^{\pi_\beta}(s_k)}d^{\pi_\beta}(s_k)\max_{\pi\in\Pi}(\hat{Q}^0 - \hat{Q}^*)(s_k,\pi)\|\\
    \leq & \gamma^k \sum_{s_k}\frac{\hat{P}(s_k|\{\pi_1...\pi_{k-1}\}, s, a)}{d^{\pi_\beta}(s_k)}\|d^{\pi_\beta}(s_k)\max_{\pi\in\Pi}(\hat{Q}^0 - \hat{Q}^*)(s_k,\pi)\|\\
    \leq & \gamma^k \sup_t\sum_{s_t}\frac{\hat{P}(s_t|\{\pi_1...\pi_{t-1}\}, s, a)}{d^{\pi_\beta}(s_t)}\sum_{s_k}\|d^{\pi_\beta}(s_k)\max_{\pi\in\Pi}(\hat{Q}^0 - \hat{Q}^*)(s_k,\pi)\|\\
    = & \gamma^k \sup_t\sum_{s_t}\frac{\hat{P}(s_t|\{\pi_1...\pi_{t-1}\}, s, a)}{d^{\pi_\beta}(s_t)}\|\triangle_0\|\\
\end{align}
where $\forall i\in[1,k], \pi_i=\arg\max_{\pi\in\Pi}(\hat{Q}^{k-i} - \hat{Q}^*)(s, \pi)$. The k-step transition distribution $\hat{P}(s_k|\{\pi_1...\pi_{k}\}, s, a)$ means that starting form $s,a$, taking policy from index $1...k$, and the final state distribution at the k-step.

The inequality $(a)$ holds because,
\begin{align}
    \|\hat{Q}^k - \hat{Q}^*\|(s,a) & = \|\hat{T}^{\Pi}\hat{Q}^{k-1} - \hat{T}^\Pi\hat{Q}^*\|(s,a)\\
    & = \gamma\|\mathbb E_{\hat{P}(s_1|s, a)}(\hat{Q}^{k-1} - \hat{Q}^*)(s_1,\pi_1)\|\\
    & \leq \gamma\|\mathbb E_{\hat{P}(s_1|s, a)}(\gamma\mathbb E_{\hat{P}(s_2|\pi_1, s_1)}(\hat{Q}^{k-2} - \hat{Q}^*)(s_2,\pi_2))\|\\
    & = \gamma^2\|\mathbb E_{\hat{P}(s_2|\pi_1, s, a)}(\hat{Q}^{k-2} - \hat{Q}^*)(s_2,\pi_2)\|\\
    &...........\\
    & = \gamma^k\|\mathbb E_{\hat{P}(s_k|\{\pi_1...\pi_{k-1}\}, s, a)}(\hat{Q}^{0} - \hat{Q}^*)(s_k,\pi_k)\|
\end{align}

Then $\forall \pi\in \Pi, s\in\mathcal{D}$, $\sup_{s'}\frac{\hat{P}(s'|s,\pi)}{d^{\pi_\beta}(s')}\leq \epsilon$, so we have, $\forall t$, $\sup_{s_t}\frac{\hat{P}(s_t|s,\{\pi_1...\pi_t\})}{d^{\pi_\beta}(s_t)}\leq\epsilon^t\leq\epsilon$, where $\pi_1...\pi_t\in\Pi$. Therefore, finally we have,
\begin{align}
    \|\hat{Q}^k - \hat{Q}^*\|(s,a)\leq \gamma^k\cdot\epsilon\cdot\|\triangle_0\|
\end{align}
Completing the proof.

\subsection{Proof of Theorem \ref{theorem_cumorl}} \label{proofoftheorem}

The proof of Theorem \ref{theorem_cumorl} is performed under Assumption \ref{assumption_bounduncertainty}.



\textbf{Theorem \ref{theorem_cumorl}.}{\it  Given an arbitrary state $s$, a conservative policy $\pi$ and a state estimator $U^{\pi}_{\mathcal D}$ based on the policy $\pi$ and dataset $\mathcal D$. Then the minimizing the ODAF term in Eq.(\ref{cumorleq}), i.e.,}
    \begin{align}
    \min_{\pi} \sum_{s'}P(s'|s, \pi)U^{\pi}_{\mathcal D}(s')\label{eq:appendixobj}
    \end{align}
{\it    is equivalent to minimizing the upper bound of the following objective,}
    \begin{align}
        \sum_{s'\notin supp(d^{\pi_\beta}(s'))}P(s'|s, \pi)\label{eq:objappendix2}
    \end{align}
 {\it  where $supp(d^{\pi_\beta}(s'))$ is the support of the dataset.}

\begin{proof}
    \begin{align}
        & \sum_{s'}P(s'|s, \pi)U^{\pi}_{\mathcal D}(s')\nonumber \\
    = & \sum_{s'\in supp(d^{\pi_\beta}(s'))}P(s'|s, \pi)U^{\pi}_{\mathcal D}(s')+ \sum_{s'\notin supp(d^{\pi_\beta}(s'))}P(s'|s, \pi)U^{\pi}_{\mathcal D}(s') \label{use_a1} \\
     & \geq \sum_{s'\notin supp(d^{\pi_\beta}(s'))}P(s'|s, \pi)U^{\pi}_{\mathcal D}(s') \label{use_a2}
    \end{align}
    
    where, by Assumption \ref{assumption_bounduncertainty}, then $\forall s'\in \mathcal D, U^{\pi}_{\mathcal D}(s')>0$, so that Eq.(\ref{use_a1}) upper bounds Eq.(\ref{use_a2}). Then, via Assumption \ref{assumption_bounduncertainty}, we have,
    
    \begin{align}
        & \sum_{s'\notin supp(d^{\pi_\beta}(s'))}P(s'|s, \pi)U^{\pi}_{\mathcal D}(s')\nonumber\\
        = & \sum_{s'\notin supp(d^{\pi_\beta}(s'))}P(s'|s, \pi)\mathbb E_{a'\sim\pi(\cdot|s')}u(s',a')\nonumber \\
        \geq & \sum_{s'\notin supp(d^{\pi_\beta}(s'))}P(s'|s, \pi)\mathbb E_{a'\sim\pi(\cdot|s')}U_{min}\nonumber \\
        = & U_{min}\cdot \sum_{s'\notin supp(d^{\pi_\beta}(s'))}P(s'|s, \pi)\nonumber
    \end{align}
    where $U_{min}$ is a constant according to $\pi$. Therefore we have,
    \begin{align}
        & \min_{\pi} U_{min}\cdot \sum_{s'\notin supp(d^{\pi_\beta}(s'))}P(s'|s, \pi)\nonumber \\
    \Leftrightarrow & \min_{\pi} \sum_{s'\notin supp(d^{\pi_\beta}(s'))}P(s'|s, \pi)\nonumber
    \end{align}
    Complete the proof.
\end{proof}





\section{Experimental Details}\label{externel_experiments}




In this section, we introduce our training details, including: 1) the hyperparameters our method use; 2) the structure of the neural networks we use: the Q-networks, inverse dynamics model network and policy network; 3) the training details of ODAF; 4) the total amount of compute and the type of resources used.

\subsection{Hyperparameters of ODAF}

In Table \ref{hpt1} and Table \ref{hpt2}, we give the hyperparameters used by ODAF to generate Table \ref{comparison} results. The $\epsilon_{odaf}$ is the perturbation scalar of a perturbation ball $B^{\epsilon_{odaf}}_s$ around state $s$ in ODAF loss and $\beta_{odaf}$ is the weight of the ODAF loss. 

\begin{table}[H]
\centering
\caption{Hyperparameters of ODAF in standard MuJoCo benchmarks.}
\begin{tabular}{llll}
\toprule
\textbf{}          & \textbf{Halfcheetah} & \textbf{Hopper} & \textbf{Walker2d} \\ \midrule
$\epsilon_{odaf}$ & 0.001                        & 0.005                  & 0.01                     \\
$\beta_{odaf}$ & 0.3                        & 0.3                  & 0.3                     \\  \bottomrule
\end{tabular}
\label{hpt1}
\end{table}

\begin{table}[H]
\centering
\caption{Hyperparameters of ODAF in adversarial attack MuJoCo benchmarks.}
\begin{tabular}{llll}
\toprule
\textbf{}          & \textbf{Halfcheetah} & \textbf{Hopper} & \textbf{Walker2d} \\ \midrule
$\epsilon_{odaf}$ & 0.05                        & 0.005                  & 0.07                     \\
$\beta_{odaf}$ & 0.3                        & 0.3                  & 0.3                     \\  \bottomrule
\end{tabular}
\label{hpt2}
\end{table}

\subsection{Neural network structures of ODAF}

In this section, we introduce the structure of the networks we use in this paper: policy network, Q network and the dynamics model network. 

The structure of the policy network and Q networks is as shown in Table \ref{policyq}, where 's\_dim' is the dimension of states and 'a\_dim' is the dimension of actions. 'h\_dim' is the dimension of the hidden layers, which is usually 256 in our experiments. The policy network is a Guassian policy and the Q networks includes ten Q function networks and ten target Q function networks.

\begin{table}[h]
    \centering
    \caption{The structure of the policy net and the Q networks.}
    \begin{tabular}{ll}
        \toprule
policy net & Q net\\
		\midrule
Linear(s\_dim, 256) & Linear(s\_dim, h\_dim)\\
Relu() & Relu()\\
Linear(h\_dim, h\_dim) & Linear(h\_dim, h\_dim)\\
Relu() & Relu()\\
Linear(h\_dim, a\_dim) & Linear(h\_dim, 1)\\
        \bottomrule
    \end{tabular}
    \label{policyq}
\end{table}

\begin{table}[h]
    \centering
    \caption{The structure of the dynamics model network.}
    \begin{tabular}{l|l}
        \toprule
\multicolumn{2}{l}{dynamics model net}\\
		\midrule
\multicolumn{2}{l}{Linear(s\_dim + a\_dim, h\_dim)} \\
\multicolumn{2}{l}{Linear(h\_dim, h\_dim)} \\
\multicolumn{2}{l}{Linear(h\_dim, h\_dim)} \\
Linear(h\_dim, z\_dim) & Linear(h\_dim, z\_dim) \\
\multicolumn{2}{l}{Linear(s\_dim + a\_dim + z\_dim, h\_dim)} \\
\multicolumn{2}{l}{Linear(h\_dim, h\_dim)}  \\
\multicolumn{2}{l}{Linear(h\_dim, s\_dim)} \\
        \bottomrule
    \end{tabular}
    \label{dmn}
\end{table}

 The structure of the dynamics network is as shown in Table \ref{dmn}, which is a conditional variational auto-encoder. 's\_dim' is the dimension of states, 'a\_dim' is the dimension of actions and 'h\_dim' is the dimension of the hidden variables. 'z\_dim' is the dimension of the Gaussian hidden variables in conditional variational auto-encoder.

\subsection{Compute resources}

We conducted all our experiments using a server equipped with one Intel Xeon Gold 5218 CPU, with 32 cores and 64 threads, and 256GB of DDR4 memory. We used a NVIDIA RTX3090 GPU with 24GB of memory for our deep learning experiments. All computations were performed using Python 3.8 and the PyTorch deep learning framework.

\bibliographystyle{plain}
\bibliography{main}

\begin{thebibliography}{10}

\bibitem{achiam2023gpt}
Josh Achiam, Steven Adler, Sandhini Agarwal, Lama Ahmad, Ilge Akkaya, Florencia~Leoni Aleman, Diogo Almeida, Janko Altenschmidt, Sam Altman, Shyamal Anadkat, et~al.
\newblock Gpt-4 technical report.
\newblock {\em arXiv preprint arXiv:2303.08774}, 2023.

\bibitem{markovrep}
Cameron Allen, Neev Parikh, Omer Gottesman, and George Konidaris.
\newblock Learning markov state abstractions for deep reinforcement learning.
\newblock In Marc'Aurelio Ranzato, Alina Beygelzimer, Yann~N. Dauphin, Percy Liang, and Jennifer~Wortman Vaughan, editors, {\em Advances in Neural Information Processing Systems 34: Annual Conference on Neural Information Processing Systems 2021, NeurIPS 2021, December 6-14, 2021, virtual}, pages 8229--8241, 2021.

\bibitem{edac}
Gaon An, Seungyong Moon, Jang{-}Hyun Kim, and Hyun~Oh Song.
\newblock Uncertainty-based offline reinforcement learning with diversified q-ensemble.
\newblock In Marc'Aurelio Ranzato, Alina Beygelzimer, Yann~N. Dauphin, Percy Liang, and Jennifer~Wortman Vaughan, editors, {\em Advances in Neural Information Processing Systems 34: Annual Conference on Neural Information Processing Systems 2021, NeurIPS 2021, December 6-14, 2021, virtual}, pages 7436--7447, 2021.

\bibitem{pbrl}
Chenjia Bai, Lingxiao Wang, Zhuoran Yang, Zhi{-}Hong Deng, Animesh Garg, Peng Liu, and Zhaoran Wang.
\newblock Pessimistic bootstrapping for uncertainty-driven offline reinforcement learning.
\newblock In {\em The Tenth International Conference on Learning Representations, {ICLR} 2022, Virtual Event, April 25-29, 2022}. OpenReview.net, 2022.

\bibitem{cheng2022adversarially}
Ching-An Cheng, Tengyang Xie, Nan Jiang, and Alekh Agarwal.
\newblock Adversarially trained actor critic for offline reinforcement learning.
\newblock In {\em International Conference on Machine Learning}, pages 3852--3878. PMLR, 2022.

\bibitem{d4rl}
Justin Fu, Aviral Kumar, Ofir Nachum, George Tucker, and Sergey Levine.
\newblock {D4RL:} datasets for deep data-driven reinforcement learning.
\newblock {\em CoRR}, abs/2004.07219, 2020.

\bibitem{td3bc}
Scott Fujimoto and Shixiang~Shane Gu.
\newblock A minimalist approach to offline reinforcement learning.
\newblock {\em Advances in neural information processing systems}, 34:20132--20145, 2021.

\bibitem{BCQ}
Scott Fujimoto, David Meger, and Doina Precup.
\newblock Off-policy deep reinforcement learning without exploration.
\newblock In Kamalika Chaudhuri and Ruslan Salakhutdinov, editors, {\em Proceedings of the 36th International Conference on Machine Learning, {ICML} 2019, 9-15 June 2019, Long Beach, California, {USA}}, volume~97 of {\em Proceedings of Machine Learning Research}, pages 2052--2062. {PMLR}, 2019.

\bibitem{ghavamzadeh2016safe}
Mohammad Ghavamzadeh, Marek Petrik, and Yinlam Chow.
\newblock Safe policy improvement by minimizing robust baseline regret.
\newblock {\em Advances in Neural Information Processing Systems}, 29, 2016.

\bibitem{sac}
Tuomas Haarnoja, Aurick Zhou, Pieter Abbeel, and Sergey Levine.
\newblock Soft actor-critic: Off-policy maximum entropy deep reinforcement learning with a stochastic actor.
\newblock In Jennifer~G. Dy and Andreas Krause, editors, {\em Proceedings of the 35th International Conference on Machine Learning, {ICML} 2018, Stockholmsm{\"{a}}ssan, Stockholm, Sweden, July 10-15, 2018}, volume~80 of {\em Proceedings of Machine Learning Research}, pages 1856--1865. {PMLR}, 2018.

\bibitem{osr}
Ke~Jiang, Jia-Yu Yao, and Xiaoyang Tan.
\newblock Recovering from out-of-sample states via inverse dynamics in offline reinforcement learning.
\newblock In {\em Thirty-seventh Conference on Neural Information Processing Systems}, 2023.

\bibitem{pessimism}
Ying Jin, Zhuoran Yang, and Zhaoran Wang.
\newblock Is pessimism provably efficient for offline rl?
\newblock In Marina Meila and Tong Zhang, editors, {\em Proceedings of the 38th International Conference on Machine Learning, {ICML} 2021, 18-24 July 2021, Virtual Event}, volume 139 of {\em Proceedings of Machine Learning Research}, pages 5084--5096. {PMLR}, 2021.

\bibitem{ispessimism}
Ying Jin, Zhuoran Yang, and Zhaoran Wang.
\newblock Is pessimism provably efficient for offline rl?
\newblock In Marina Meila and Tong Zhang, editors, {\em Proceedings of the 38th International Conference on Machine Learning, {ICML} 2021, 18-24 July 2021, Virtual Event}, volume 139 of {\em Proceedings of Machine Learning Research}, pages 5084--5096. {PMLR}, 2021.

\bibitem{cem}
Dmitry Kalashnikov, Alex Irpan, Peter Pastor, Julian Ibarz, Alexander Herzog, Eric Jang, Deirdre Quillen, Ethan Holly, Mrinal Kalakrishnan, Vincent Vanhoucke, et~al.
\newblock Scalable deep reinforcement learning for vision-based robotic manipulation.
\newblock In {\em Conference on robot learning}, pages 651--673. PMLR, 2018.

\bibitem{cvae}
Diederik~P. Kingma and Max Welling.
\newblock Auto-encoding variational bayes.
\newblock In Yoshua Bengio and Yann LeCun, editors, {\em 2nd International Conference on Learning Representations, {ICLR} 2014, Banff, AB, Canada, April 14-16, 2014, Conference Track Proceedings}, 2014.

\bibitem{iql}
Ilya Kostrikov, Ashvin Nair, and Sergey Levine.
\newblock Offline reinforcement learning with implicit q-learning.
\newblock In {\em The Tenth International Conference on Learning Representations, {ICLR} 2022, Virtual Event, April 25-29, 2022}. OpenReview.net, 2022.

\bibitem{cql}
Aviral Kumar, Aurick Zhou, George Tucker, and Sergey Levine.
\newblock Conservative q-learning for offline reinforcement learning.
\newblock In Hugo Larochelle, Marc'Aurelio Ranzato, Raia Hadsell, Maria{-}Florina Balcan, and Hsuan{-}Tien Lin, editors, {\em Advances in Neural Information Processing Systems 33: Annual Conference on Neural Information Processing Systems 2020, NeurIPS 2020, December 6-12, 2020, virtual}, 2020.

\bibitem{li2024temporal}
Ruoting Li, Joseph~K Agor, and Osman~Y {\"O}zalt{\i}n.
\newblock Temporal pattern mining for knowledge discovery in the early prediction of septic shock.
\newblock {\em Pattern Recognition}, 151:110436, 2024.

\bibitem{li2025novel}
Tengbiao Li and Junsheng Qiao.
\newblock A novel q-rung orthopair fuzzy magdm method for healthcare waste treatment based on three-way decisions.
\newblock {\em Pattern Recognition}, 157:110867, 2025.

\bibitem{li2022alleviating}
Yao Li, YuHui Wang, YaoZhong Gan, and XiaoYang Tan.
\newblock Alleviating the estimation bias of deep deterministic policy gradient via co-regularization.
\newblock {\em Pattern Recognition}, 131:108872, 2022.

\bibitem{li2023self}
Yao Li, YuHui Wang, and XiaoYang Tan.
\newblock Self-imitation guided goal-conditioned reinforcement learning.
\newblock {\em Pattern Recognition}, 144:109845, 2023.

\bibitem{mao2024offline}
Yixiu Mao, Cheems Wang, Chen Chen, Yun Qu, and Xiangyang Ji.
\newblock Offline reinforcement learning with ood state correction and ood action suppression.
\newblock {\em arXiv preprint arXiv:2410.19400}, 2024.

\bibitem{svr}
Yixiu Mao, Hongchang Zhang, Chen Chen, Yi~Xu, and Xiangyang Ji.
\newblock Supported value regularization for offline reinforcement learning.
\newblock In {\em Thirty-seventh Conference on Neural Information Processing Systems}, 2023.

\bibitem{mnih2015human}
Volodymyr Mnih, Koray Kavukcuoglu, David Silver, Andrei~A Rusu, Joel Veness, Marc~G Bellemare, Alex Graves, Martin Riedmiller, Andreas~K Fidjeland, Georg Ostrovski, et~al.
\newblock Human-level control through deep reinforcement learning.
\newblock {\em nature}, 518(7540):529--533, 2015.

\bibitem{pang2025qfae}
Teng Pang, Guoqiang Wu, Yan Zhang, Bingzheng Wang, and Yilong Yin.
\newblock Qfae: Q-function guided action exploration for offline deep reinforcement learning.
\newblock {\em Pattern Recognition}, 158:111032, 2025.

\bibitem{peng2017deeploco}
Xue~Bin Peng, Glen Berseth, KangKang Yin, and Michiel Van De~Panne.
\newblock Deeploco: Dynamic locomotion skills using hierarchical deep reinforcement learning.
\newblock {\em Acm transactions on graphics (tog)}, 36(4):1--13, 2017.

\bibitem{silver2017mastering}
David Silver, Julian Schrittwieser, Karen Simonyan, Ioannis Antonoglou, Aja Huang, Arthur Guez, Thomas Hubert, Lucas Baker, Matthew Lai, Adrian Bolton, et~al.
\newblock Mastering the game of go without human knowledge.
\newblock {\em nature}, 550(7676):354--359, 2017.

\bibitem{touvron2023llama}
Hugo Touvron, Louis Martin, Kevin Stone, Peter Albert, Amjad Almahairi, Yasmine Babaei, Nikolay Bashlykov, Soumya Batra, Prajjwal Bhargava, Shruti Bhosale, et~al.
\newblock Llama 2: Open foundation and fine-tuned chat models.
\newblock {\em arXiv preprint arXiv:2307.09288}, 2023.

\bibitem{ql}
Christopher J. C.~H. Watkins and Peter Dayan.
\newblock Technical note q-learning.
\newblock {\em Mach. Learn.}, 8:279--292, 1992.

\bibitem{SPOT}
Jialong Wu, Haixu Wu, Zihan Qiu, Jianmin Wang, and Mingsheng Long.
\newblock Supported policy optimization for offline reinforcement learning.
\newblock In Sanmi Koyejo, S.~Mohamed, A.~Agarwal, Danielle Belgrave, K.~Cho, and A.~Oh, editors, {\em Advances in Neural Information Processing Systems 35: Annual Conference on Neural Information Processing Systems 2022, NeurIPS 2022, New Orleans, LA, USA, November 28 - December 9, 2022}, 2022.

\bibitem{brac}
Yifan Wu, George Tucker, and Ofir Nachum.
\newblock Behavior regularized offline reinforcement learning.
\newblock {\em arXiv preprint arXiv:1911.11361}, 2019.

\bibitem{BEAR}
Yifan Wu, George Tucker, and Ofir Nachum.
\newblock Behavior regularized offline reinforcement learning.
\newblock {\em arXiv preprint arXiv:1911.11361}, 2019.

\bibitem{xie2021bellman}
Tengyang Xie, Ching-An Cheng, Nan Jiang, Paul Mineiro, and Alekh Agarwal.
\newblock Bellman-consistent pessimism for offline reinforcement learning.
\newblock {\em Advances in neural information processing systems}, 34:6683--6694, 2021.

\bibitem{rorl}
Rui Yang, Chenjia Bai, Xiaoteng Ma, Zhaoran Wang, Chongjie Zhang, and Lei Han.
\newblock {RORL:} robust offline reinforcement learning via conservative smoothing.
\newblock {\em CoRR}, abs/2206.02829, 2022.

\bibitem{yu2021combo}
Tianhe Yu, Aviral Kumar, Rafael Rafailov, Aravind Rajeswaran, Sergey Levine, and Chelsea Finn.
\newblock Combo: Conservative offline model-based policy optimization.
\newblock {\em Advances in neural information processing systems}, 34:28954--28967, 2021.

\bibitem{mopo}
Tianhe Yu, Garrett Thomas, Lantao Yu, Stefano Ermon, James~Y. Zou, Sergey Levine, Chelsea Finn, and Tengyu Ma.
\newblock {MOPO:} model-based offline policy optimization.
\newblock In Hugo Larochelle, Marc'Aurelio Ranzato, Raia Hadsell, Maria{-}Florina Balcan, and Hsuan{-}Tien Lin, editors, {\em Advances in Neural Information Processing Systems 33: Annual Conference on Neural Information Processing Systems 2020, NeurIPS 2020, December 6-12, 2020, virtual}, 2020.

\bibitem{sdc}
Hongchang Zhang, Jianzhun Shao, Yuhang Jiang, Shuncheng He, Guanwen Zhang, and Xiangyang Ji.
\newblock State deviation correction for offline reinforcement learning.
\newblock In {\em Thirty-Sixth {AAAI} Conference on Artificial Intelligence, {AAAI} 2022}, pages 9022--9030. {AAAI} Press, 2022.

\bibitem{zhangzhe}
Zhe Zhang and Xiaoyang Tan.
\newblock An implicit trust region approach to behavior regularized offline reinforcement learning.
\newblock In {\em Proceedings of the AAAI Conference on Artificial Intelligence}, volume~38, pages 16944--16952, 2024.

\bibitem{zhou2023free}
Zhaoyi Zhou, Chuning Zhu, Runlong Zhou, Qiwen Cui, Abhishek Gupta, and Simon~Shaolei Du.
\newblock Free from bellman completeness: Trajectory stitching via model-based return-conditioned supervised learning.
\newblock {\em arXiv preprint arXiv:2310.19308}, 2023.

\end{thebibliography}

\end{document}